\documentclass{article}
\usepackage{ijcai22}

\usepackage{times}
\usepackage{soul}
\usepackage{url}
\usepackage[hidelinks]{hyperref}
\usepackage[utf8]{inputenc}
\usepackage[small]{caption}
\usepackage{graphicx}
\usepackage{amsmath}
\usepackage{amsthm}
\usepackage{booktabs}
\usepackage{algorithm}
\usepackage{algorithmic}
\usepackage{amsthm}
\usepackage{multirow}
\usepackage{amsmath}
\usepackage{tabularx}
\usepackage{booktabs}
\usepackage{enumitem}
\usepackage{bbm}
\usepackage{xcolor}
\usepackage{balance}
\usepackage[flushleft]{threeparttable}

\newtheorem{problem}{Problem}

\DeclareMathOperator*{\argmin}{arg\,min}
\usepackage{color, colortbl}
\definecolor{Gray}{gray}{0.9}

\title{Few-Shot Learning on Graphs}

\author{{
Chuxu Zhang$^1$ 
\and
Kaize Ding$^{2}$
\and
Jundong Li$^{3}$
\and
\\
Xiangliang Zhang$^{4}$
\and
Yanfang Ye$^{4}$
\and
Nitesh V. Chawla$^{4}$
\and
Huan Liu$^{2}$
}
\affiliations {
$^1$Brandeis University, Waltham, MA, USA\\
$^2$Arizona State University, Tempe, Arizona, USA\\
$^3$University of Virginia, Charlottesville, Virginia, USA\\
$^4$University of Notre Dame, Notre Dame, Indiana, USA
}\\
\emails
{chuxuzhang@brandeis.edu, \{kaize.ding, huan.liu\}@asu.edu, jl6qk@virginia.edu, \\
\{xzhang33, yye7, nchawla\}@nd.edu}}

\begin{document}

\maketitle

\begin{abstract}
Graph representation learning has attracted tremendous attention due to its remarkable performance in many real-world applications. However, prevailing supervised graph representation learning models for specific tasks often suffer from label sparsity issue as data labeling is always time and resource consuming. In light of this, \textbf{\underline{f}}ew-\textbf{\underline{s}}hot \textbf{\underline{l}}earning on \textbf{\underline{g}}raphs (FSLG), which combines the strengths of graph representation learning and few-shot learning together, has been proposed to tackle the performance degradation in face of limited annotated data  challenge. There have been many studies working on FSLG recently. In this paper, we comprehensively survey these work in the form of a series of methods and applications. Specifically, we first introduce FSLG challenges and bases, then categorize and summarize existing work of FSLG in terms of three major graph mining tasks at different granularity levels, i.e., node, edge, and graph. Finally, we share our thoughts on some future research directions of FSLG. The authors of this survey have contributed significantly to the AI literature on FSLG over the last few years. 

\end{abstract}

\section{Introduction}
Many real-world systems can be modeled as graphs, representing nodes (entities) interconnected by edges (relations) as well as attributes in nodes and edges. Traditional graph mining algorithms usually require both domain understanding and exploratory search when doing feature engineering of a specific task. Accordingly,
graph representation learning techniques (GRL)~\cite{zhang2020deep}, e.g., graph neural networks, have been proposed to automatically generate graph representation for various downstream applications across different domains, such as social/information systems~\cite{perozzi2014deepwalk,grover2016node2vec,kipf2016semi}, bioinformatics/cheminformatics~\cite{jin2017predicting,hao2020asgn}, and recommender systems~\cite{ying2018graph,fan2019graph}. 

\begin{figure}[t]
	\centering
	\includegraphics[width=0.9\columnwidth]{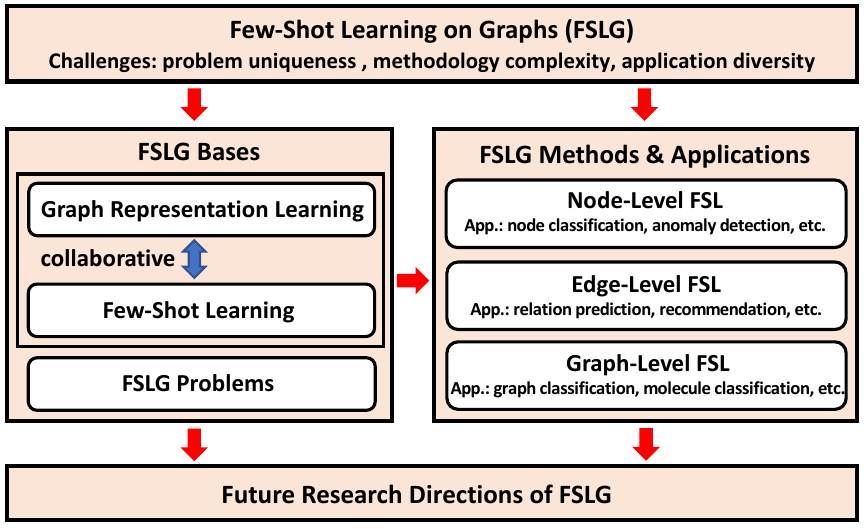}
	\caption{The framework of this survey.}
	\label{fig:intro}
\end{figure}

Developing a powerful supervised GRL model for a specific downstream task often requires abundant annotated samples. However, lacking labeled data is common in real situations due to the expensive labeling cost~\cite{ding2022meta}. For example, molecular property testing for therapeutic activity spends a lot of money on human-laboring and experimental resource~\cite{guo2021few}; recommender systems face cold-start problem for new users (items) coming to the system~\cite{lu2020meta}. This calls for a new GRL paradigm that can effectively learn graph representation for various graph mining tasks with the limited labeled data constraint. Fortunately, few-shot learning (FSL)~\cite{wang2020generalizing} has emerged to alleviate the dependence on labeled data and learn effective data representation for image/vision~\cite{vinyals2016matching,snell2017prototypical}, language/text~\cite{yu2018diverse,hu2019few}, and robotics~\cite{finn2017model,duan2017one}, using a small amount of labels. 

Therefore, \textbf{\underline{f}}ew-\textbf{\underline{s}}hot \textbf{\underline{l}}earning on \textbf{\underline{g}}raphs (FSLG), which generally combines the advantages of GRL and FSL, has become a promising research topic and attracted increasing attention from the AI community. Consequently, there have been many FSLG studies that span a variety of methods and applications in recent years.  In this paper, we provide a comprehensive and systematic review of FSLG. Figure~\ref{fig:intro} illustrates the framework of this survey. To be specific, we first analyze the key challenges of FSLG. Then we present FSLG bases consisting of two collaborative machine learning techniques: graph representation learning and few-shot learning, and introduce three typical FSLG problems. Next, we summarize existing FSLG methods and applications that are categorized into three parts corresponding to three typical graph mining tasks at different granularity levels: node-level FSL, edge-level FSL, and graph-level FSL. For each part, we cover two major lines of work: metric-based methods and optimization-based methods, and further discuss related applications. We also provide a comprehensive summary table that lists the representative FSLG work with their open-source codes/datasets. Finally, we discuss open problems and pressing issues as future research directions of FSLG. To summarize, our contributions of this work are:
\begin{itemize}[leftmargin=*]\setlength{\itemsep}{0pt}
\item We analyze the key challenges of FSLG from perspectives of problem, methodology, and applications. 
\item We comprehensively survey existing studies of FSLG  by systematically categorizing them into three parts according to different granularity levels of graph mining tasks.
\item We discuss some future research directions of FSLG, which may shed light on the development of the AI community. 
\end{itemize}

\section{FSLG Challenges and Bases}
FSLG, as a new and promising research topic in the AI community, is non-trivial and faces the following key challenges:
\begin{itemize}[leftmargin=*]\setlength{\itemsep}{0pt}
\item \textbf{The uniqueness of FSLG problem.} Unlike the general GRL problem, FSLG faces the challenge of limited labels. Besides, FSLG is different from FSL on image or text as graph lies in a non-Euclidean space and has more complex characteristics. Therefore, the uniqueness of FSLG problem requires handling both graph data complexity (e.g., irregularity, heterogeneity) and few-shot task difficulty (e.g., task diversity, inductive bias). 
\item \textbf{The complexity of methodology.} 
Before FSLG emerges, there are extensive studies targeting GRL and FSL challenges independently while none of them are able to address both of them jointly. Thus, it is essential yet difficult to develop FSLG methods that consist of both GRL and FSL modules, and combine them for solving the FSLG problem in a collaborative and effective manner. 
\item \textbf{The diversity of graph mining applications.} Graph mining tasks regarding different applications are diverse, ranging from node-level to edge-level to graph-level tasks. They require different settings, objectives, constraints, and domain knowledge. Hence, it is not easy to develop a customized FSLG method for the target application. 
\end{itemize}

FSLG methods and applications reviewed in Section~\ref{sec:method and application} aim to tackle at least one of the above three challenges. In the following of this section, we introduce FSLG bases that include two learning techniques collaboratively used in FSLG, and three FSLG problems that share the same setting.

\subsection{Graph Representation Learning}
\label{sec:grl}
The purpose of graph representation learning (GRL)~\cite{zhang2020deep} is to automate the discovery of meaningful vector representation of nodes, edges, or the whole graph for various downstream graph mining applications. Existing GRL approaches generally fall into three groups: (1) network embedding models~\cite{perozzi2014deepwalk,grover2016node2vec,dong2017metapath2vec} that capture graph structure information by preserving proximities among contextual nodes; (2) graph neural networks (GNNs)~\cite{kipf2016semi,velivckovic2018graph,zhang2019hetgnn} that learn node embedding by aggregating neighbors' feature information; and (3) knowledge graph embedding methods~\cite{bordes2013translating,socher2013reasoning,dettmers2018convolutional} that construct graph as a collection of fact triplets and learn node and edge (a.k.a. entity and relation) embedding through modeling the acceptability score of each fact triplet. 

GNNs, as the current state-of-the-art in GRL, are most commonly utilized to build the GRL backbone of a FSLG method. Specifically, a graph is represented as $G = (V, E, X)$, where $V$ is the set of nodes, $E \subseteq  V \times V$ is the set of edges, and $X$ is the set of node (and edge) attributes. GNNs learn node embedding via message-passing framework: 
\begin{equation}
    h_v^{(l+1)} = \mathrm{COM}\left (h_v^{(l)}, \left[\mathrm{AGG}\left(\left\{h_{u}^{(l)}~|~\forall u \in \mathcal{N}_v\right\}\right)\right]\right), 
\label{eq:gnn}
\end{equation}
where $h_v^{(l)}$ denotes embedding of node $v$ at $l$-th GNN layer;  $\mathcal{N}_v$ is the neighbor set of node $v$; $\mathrm{AGG}(\cdot)$ and $\mathrm{COM}(\cdot)$ are neighbor aggregation and combination functions, respectively; $h_v^{(0)}$ is initialized with node attribute $X_{v}$. Furthermore, the whole graph embedding can be computed as:
\begin{equation}
    h_G^{(l)} = \textrm{READOUT}\left\{h_{v}^{(l)}~|~\forall v \in V\right\}, 
\label{eq:gnn-graph}
\end{equation}
where the READOUT function can be a simple permutation invariant function such as summation or a more sophisticated graph-level pooling function~\cite{ying2018hierarchical}. 

\subsection{Few-Shot Learning}
\label{sec:fsl}
Few-shot learning (FSL)~\cite{wang2020generalizing}
aims to learn generalized experiences from existing tasks to form transferable prior knowledge for new tasks with limited labeled data. It commonly adopts a meta-learning framework~\cite{hospedales2020meta} which performs episodic learning to train and optimize the model. Specifically, given a set of tasks $\mathcal{T}$ and their data, in the meta-training phase, the objective is to learn parameters $\theta^{*}$ that work effectively across all tasks in $\mathcal{T}$:
\begin{equation}
    \theta^{*} = \argmin_{\theta} \sum_{\mathcal{T}_i \sim p(\mathcal{T})} \mathcal{L}\left(\mathcal{D}_{\mathcal{T}_i}, \theta\right), 
\label{eq:fsl-1}
\end{equation}
where $p(\mathcal{T})$ denotes task distribution; $\mathcal{D}_{\mathcal{T}_i}$ is the data of task $\mathcal{T}_i$; $\mathcal{L}$ is the loss function for a downstream application. In the meta-testing phase, $\theta^{*}$ is taken as the initialized parameters (meta-knowledge) that are further quickly adapted to a new task $\mathcal{T}_j$: 
\begin{equation}
    \theta^{**} = \mathcal{L}\left(\mathcal{D}_{\mathcal{T}_j}, \theta^{*}\right).
\label{eq:fsl-2}
\end{equation}
Note that, $\mathcal{D}_{\mathcal{T}_j}$ only contains limited labeled data. Notable FSL methods used in FSLG models generally fall into two categories: (1) metric-based methods~\cite{vinyals2016matching,snell2017prototypical,sung2018learning} that learn task-specific similarity metric between query data and support set data; and (2) optimization-based methods~\cite{ravi2016optimization,finn2017model,finn2018probabilistic} that learn well initialized base-learner which can quickly adapt to a new few-shot task with gradient computation. 

\subsection{FSLG Problems}
\label{sec:problem}
Different FSLG problems share the same setting of FSL. In specific, let $\mathcal{C}$ denote the entire classes set of the whole dataset, which can be further divided into two categories: base classes set $\mathcal{C}_{base}$ of the training data and new (novel) classes set $\mathcal{C}_{novel}$ of the testing data, where $\mathcal{C} = \mathcal{C}_{base} \cup  \mathcal{C}_{novel}$ and $\mathcal{C}_{base} \cap \mathcal{C}_{novel} = \emptyset$. Generally, the number of labels is abundant in $\mathcal{C}_{base}$ while scarce in $\mathcal{C}_{novel}$. 
Here, we introduce three typical FSLG problems corresponding to three graph mining tasks.

\begin{problem}
\textbf{Few-Shot Node Classification}. Given a graph, the problem is to develop a machine learning model such that after training on labeled nodes in $\mathcal{C}_{base}$, the model can accurately predict labels for nodes (query set) in $\mathcal{C}_{novel}$ with only a limited number of labeled nodes (support set).  
\end{problem}

\begin{problem}
\textbf{Few-Shot Relation Prediction}. Given a graph, the problem is to develop a machine learning model such that after training on node pairs of relations in $\mathcal{C}_{base}$, the model can accurately predict unknown node pairs for relations (query set) in $\mathcal{C}_{novel}$ with only a limited number of known node pairs (support set).  
\end{problem}

\begin{problem}
\textbf{Few-Shot Graph Classification}. Given a set of graphs, the problem is to develop a machine learning model such that after training on labeled graphs in $\mathcal{C}_{base}$, the model can accurately predict labels for graphs (query set) in $\mathcal{C}_{novel}$ with only a limited number of labeled graphs (support set).  
\end{problem}

The FSL setting is applied to different FSLG problems by setting the class meaning. Specifically, each class corresponds to a node label for the node classification problem, a relation type for the relation prediction problem, and a graph label for the graph classification problem. Note that, if the support set contains exactly $K$ nodes for each of $N$ classes from $\mathcal{C}_{novel}$ and the query set is sampled from these $N$ classes, the problem is called $N$-way $K$-shot problem. Besides the above three typical problems, there are other FSLG problems, such as few-shot anomaly detection and few-shot recommendation on graphs. 

\section{FSLG Methods and Applications}
\label{sec:method and application}
In general, FSLG combines the strengths of GRL (Section~\ref{sec:grl}) and FSL (Section~\ref{sec:fsl}) together for various applications. In this section, we comprehensively review current methods and applications of FSLG by systematically categorizing them into three parts corresponding to three problems in Section~\ref{sec:problem}: node-level FSL, edge-level FSL, and graph-level FSL. Note that, most of FSLG methods adopt two typical FSL techniques: metric-based methods and optimization-based methods. Accordingly, methods of each part are summarized into two groups, depending on which FSL technique they rely on, followed by the discussion of related applications. At the end, a list of representative FSLG studies with their open-source codes/datasets are shown in Table~\ref{tab:summary}. 

\subsection{Node-level FSL}
Node is the fundamental unit of which graphs are formed. Node-level learning not only facilitates a variety of node-based applications (e.g., node classification and anomaly detection), but also lays the groundwork for further edge-level and graph-level applications. However, it is often difficult to collect node labels in real practice. For example, obtaining function labels for proteins in the interactome (i.e., protein-protein network) is a time and labor-consuming task even for experienced experts~\cite{wang2020graph}; a significant number of research venues (labels) in DBLP data (i.e., academic network) have few labeled publications~\cite{ding2020graph}. In light of this, there have been many methods proposed to solve node-level problems with limited labeled data. We review existing studies of node-level FSL in this part. 
\\
\textbf{Metric-based Method.}
Basically, Metric-based Node-level FSL (MN-FSL) adopts the idea of Prototypical Network (ProNet)~\cite{snell2017prototypical} which is a simple yet effective FSL framework. Specifically, MN-FSL first applies a GNN encoder to learn node embedding, and then generates the prototype of each node class by computing the mean of support nodes' embeddings. Finally, MN-FSL classifies query nodes by calculating their Euclidean distances of embeddings to different class prototypes. 
By incorporating node embedding generated by GNN into ProNet, MN-FSL is able to tackle the challenges of few-shot node classification problem. Built on the basic model, a number of variants of MN-FSL have been proposed.
Specifically, GFL~\cite{yao2020graph} designs graph-structured prototype to capture the relation structure of support samples, which is further tailored by a graph representation gate to include the whole graph information. Considering the fact that each node has a different significance in graph, GPN~\cite{ding2020graph} introduces node importance with self-attention mechanism, then computes the weighted summation of support nodes' embeddings as the refined prototype of each class. In addition, MetaTNE~\cite{lan2020node} leverages an embedding transformation function to map the task-agnostic node representation to the task-specific ones, exploiting the complex and multifaceted relationships between nodes.
More recently, HAG-Meta~\cite{tan2021graph} leverages both node-level attention and task-level attention to improve class prototype computation, which is further trained by an incremental learning paradigm. 
\\
\textbf{Optimization-based Method.}
In general, Optimization-based Node-level FSL (ON-FSL) is developed based on model-agnostic meta-learning (MAML)~\cite{finn2017model}. To be specific, ON-FSL utilizes a GNN encoder to learn node embedding for node classification. In the optimization stage, ON-FSL first updates task-specific parameters for each node class, and then accumulates all task-specific classification losses to update task-agnostic parameters. Finally, the optimized task-agnostic parameters shared by base classes are further quickly adapted to predict labels of nodes in new classes by gradient updates over a few labeled nodes. By optimizing GNN with MAML, ON-FSL learns prior experiences (meta-knowledge) across base classes for fast adaption over new classes and addresses the challenge of few-shot node classification problem. In particular, Meta-GNN~\cite{zhou2019meta} is the first work that combines GNN and MAML for node classification. Since then, there have been some improved work. Considering that feature distribution may vary across different sampled tasks, AMM-GNN~\cite{wang2020graph} introduces an attribute-level attention mechanism to better capture the unique property of each meta-learning task. In addition, G-META~\cite{huang2020graph} theoretically justifies that the evidence for a prediction can be found in the local subgraph surrounding the target node and leverages subgraph to learn node embedding. It further combines both ProNet and MAML for model optimization. Moreover,  RALE~\cite{liu2021relative} captures both task-level and graph-level dependencies to improve meta-knowledge transfer process by assigning node locations on the graph. 

In addition to the above mentioned work, there are some other studies~\cite{zhao2021multi,zhuang2021hinfshot,liu2021tail,zhang2022hg-meta} related to node-level FSL. For example, HG-Meta~\cite{zhang2022hg-meta} proposes to address few-shot node classification on heterogeneous graphs by modeling both graph structure heterogeneity and task diversity. 
In addition, unlike the aforementioned work that focus on predicting labels of nodes in new classes using few labeled samples, models that target classifying nodes with few links (tail nodes) have also been developed~\cite{liu2021tail}. 
\\
\textbf{Application.} Besides node classification on various types of graphs (e.g., social network, academic graph, and biological network), node-level FSL has been applied to some other applications. Due to the scarcity of outlier data, it is natural to develop FSLG methods to detect anomalies on graphs. In particular, Meta-GDN~\cite{ding2021few} detects network anomaly through the FSL framework augmented with a graph deviation network. Furthermore, FSLG models that incorporate domain knowledge to solve domain-specific anomaly detection problems have also been proposed. For example, Meta-AHIN~\cite{qian2021adapting} incorporates malware-related attributes and information into GNN and MAML for malicious repository detection on social coding platforms (e.g., Github). Similarly,  MetaHG~\cite{qian2021distilling} leverages drug-related features and knowledge to the joint model of GNN and MAML for illicit drug trafficker detection on social media (e.g., Instagram).  

\begin{table*}[t]
\centering
\resizebox{1\textwidth}{!}{
\begin{tabular} {c|c|c|c|c|c}
\toprule
Method & Learning Task & Learning Approach & Characteristic & Venue & Code/Data Link \\
\midrule
\rowcolor{Gray}
\multicolumn{6}{c}{Node-level FSL}\\
\midrule
GFL$^{[1]}$ & Node classification & ProNet & Graph structured prototype  & AAAI'20 & \url{https://shorturl.at/jquCS} \\
\midrule
GPN$^{[2]}$ & Node classification & ProNet & Node importance  & CIKM'20 & \url{https://shorturl.at/cxG16} \\
\midrule
MetaTNE$^{[3]}$ & Node classification & ProNet &  Embedding transformation & NeurIPS'20 & \url{https://shorturl.at/oK245}\\
\midrule
Meta-GNN$^{[4]}$ & Node classification & MAML & Basic model & CIKM'19 &  \url{https://shorturl.at/hxLP2} \\
\midrule
G-META$^{[5]}$ & Node classification& ProNet+MAML &  Local subgraph & NeurIPS'20 & \url{https://shorturl.at/zDJKL}\\
\midrule
RALE$^{[6]}$ & Node classification & MAML & Task dependency  & AAAI'21 & \url{https://shorturl.at/bsvDQ} \\
\midrule
Meta-GDN$^{[7]}$ & Anomaly detection & MAML &  Graph deviation network & WWW'21 & \url{https://shorturl.at/izQ79}\\
\midrule
MetaHG$^{[8]}$ & Anomaly detection & MAML &  Domain knowledge & NeurIPS'21 & \url{https://shorturl.at/rJV28}\\
\midrule
\rowcolor{Gray}
\multicolumn{6}{c}{Edge-level FSL}\\
\midrule
GMatching$^{[9]}$ & Relation prediction & MatchNet &  LSTM  matching processor & EMNLP'18 & \url{https://shorturl.at/vDH13}\\
\midrule
FSRL$^{[10]}$ & Relation prediction & MatchNet & Support data aggregation & AAAI'20 &  \url{https://shorturl.at/otAI4}\\
\midrule
FAAN$^{[11]}$ & Relation prediction & MatchNet & Adaptive matching & EMNLP'20 & \url{https://shorturl.at/iAGW3}\\
\midrule
GEN$^{[12]}$ & Relation prediction & MatchNet & Inductive prediction & NeurIPS'20 & \url{https://shorturl.at/esN17}\\
\midrule
REFORM$^{[13]}$ & Relation prediction & MatchNet &  Error mitigation & CIKM'21 & \url{https://shorturl.at/mpO67}\\
\midrule
MetaR$^{[14]}$ & Relation prediction & TransNet &  Relation meta & EMNLP'19 & \url{https://shorturl.at/wAGIJ}\\
\midrule
GANA$^{[15]}$ & Relation prediction & TransNet &  Refined relation meta & SIGIR'21 & \url{https://shorturl.at/mpvB3}\\
\midrule
Meta-KGR$^{[16]}$ & Multi-hop relation prediction & MAML &  Basic model & EMNLP'19 &  \url{https://shorturl.at/bmrFP}\\
\midrule
FIRE$^{[17]}$ & Multi-hop relation prediction & MAML & Space pruning & EMNLP'20 &  \url{https://shorturl.at/suwB6}\\
\midrule
ADK-KG$^{[18]}$ & Multi-hop relation prediction & MAML & Text-enhanced embedding & SDM'22 &  \url{https://shorturl.at/imzJK}\\
\midrule
\rowcolor{Gray}
\multicolumn{6}{c}{Graph-level FSL}\\
\midrule
SuperClass$^{[19]}$ & Graph classification & ProNet &  Super classes  & ICLR'20 & \url{https://shorturl.at/yPV07}\\
\midrule
AS-MAML$^{[20]}$ & Graph classification & MAML & Adaptation Controller & CIKM'20 &  \url{https://shorturl.at/svE49}\\
\midrule
Meta-MGNN$^{[21]}$ & Molecule classification & MAML & Task weight & WWW'21 &  \url{https://shorturl.at/stxAP}\\
\midrule
Pre-PAR$^{[22]}$ & Molecule classification & MAML & Property-aware embedding  & NeurIPS'21 &  \url{https://shorturl.at/sAST4}\\
\toprule
\end{tabular}
}
\begin{tablenotes}
\item 
\small{Note: $^{[1]}$\cite{yao2020graph}; $^{[2]}$\cite{ding2020graph}; $^{[3]}$\cite{lan2020node}; $^{[4]}$\cite{zhou2019meta}; $^{[5]}$\cite{huang2020graph}; $^{[6]}$\cite{liu2021relative}; $^{[7]}$\cite{ding2021few};  $^{[8]}$\cite{qian2021distilling};  $^{[9]}$\cite{xiong2018one}; $^{[10]}$\cite{zhang2020few};  $^{[11]}$\cite{sheng2020adaptive};  $^{[12]}$\cite{baek2020learning};  $^{[13]}$\cite{wang2021reform}; $^{[14]}$\cite{chen2019meta}; $^{[15]}$\cite{niu2021relational}; $^{[16]}$\cite{lv2019adapting}; $^{[17]}$\cite{zhang2020few2};$^{[18]}$\cite{zhang2022hg-adk-kg};
$^{[19]}$\cite{chauhan2019few};$^{[20]}$\cite{ma2020adaptive};
$^{[21]}$\cite{guo2021few};$^{[22]}$\cite{wang2021property}.}
\end{tablenotes}
\caption{A list of representative FSLG methods with open-source code/data.} 
\label{tab:summary}
\end{table*}

\subsection{Edge-level FSL}

Edges explicitly interconnect nodes on a graph and many applications such as relation prediction and recommendation are relied on  edge-level graph learning. However, scarcity issue of relation is prevalent in different real situations: a large portion of semantic relations only appear a few times in knowledge bases~\cite{xiong2018one}; E-commerce online platforms face cold-start problem from both user and item sides~\cite{lu2020meta}; relationships in biological interaction networks can be noisy and sparse~\cite{bose2019meta}. Therefore, researchers have been  motivated to propose extensive work to solve edge-level problems with limited labels, as summarized below.
\\
\textbf{Metric-based Method.} In general, existing Metric-based Edge-level FSL (ME-FSL) models rely on on either Matching Network (MatchNet)~\cite{vinyals2016matching} - a popular framework of FSL, or Translation Network (TransNet) - a typical knowledge graph embedding method~\cite{bordes2013translating}. For ME-FSL built on MatchNet, it first applies a GNN encoder to learn node (entity) embedding, then computes aggregated embedding of each relation by aggregating embeddings of node pairs in reference data (support set). Finally, the matching score between embeddings of relation and query data is used to determine the acceptability of each query sample. Based on this idea, GMatching~\cite{xiong2018one} is firstly proposed to solve one-shot relation prediction problem on knowledge graphs. Later, FSRL~\cite{zhang2020few} extends GMatching to few-shot scenario by attentively aggregating all support samples of each relation and improving node embedding formulation with a heterogeneous neighbor aggregator. FAAN~\cite{sheng2020adaptive} obtains further improvement over FSRL by designing an adaptive attentional network to learn adaptive node and reference representations. Furthermore, GEN~\cite{baek2020learning} investigates a more challenging out-of-graph scenario for relation prediction between unseen nodes or between seen and unseen nodes. More recently, REFORM~\cite{wang2021reform} designs an error mitigation module to alleviate the negative impact of errors incorporated into knowledge graph construction. For ME-FSL built on TransNet, it first computes relation meta by aggregating embeddings of support node pairs, and then leverages the relation meta to model correlation of query pairs using the objective loss of TransE~\cite{bordes2013translating}. Two recent models that adopt this idea are MetaR~\cite{chen2019meta} and GANA~\cite{niu2021relational}. MetaR computes relation meta by averaging all node pair-specific relation meta and performs rapid update on it for relation prediction on knowledge graphs. Furthermore, GANA extends MetaR by refining node embedding and relation meta computation with an attention mechanism and a LSTM aggregator, respectively.
In summary, by incorporating node
embedding encoded by GNN into MatchNet or TransNet, ME-FSL is able to address the challenges of few-shot relation prediction problem.
\\
\textbf{Optimization-based Method.} 
Similar to ON-FSL, Optimization-based Edge-level FSL (OE-FSL) relies on MAML for model optimization. In other words, OE-FSL tackles the challenge of few-shot relation prediction problem by optimizing GNN with MAML.
In particular, Meta-KGR~\cite{lv2019adapting} is the first work for few-shot multi-hop relation prediction on knowledge graphs. Specifically, Meta-KGR introduces a reinforcement learning framework to model multi-hop reasoning process, where the search path is encoded by a recurrent neural network. It then adopts MAML to learn effective meta parameters from high-frequency relations that could quickly adapt to few-shot relations. Later, FIRE~\cite{zhang2020few} extends Meta-KGR with a heterogeneous neighbor aggregator and a search space pruning strategy. More recently, ADK-KG~\cite{zhang2022hg-adk-kg} further improves FIRE by enhancing neighbor aggregator with node text content and augmenting MAML with task weight. 

In addition to the studies discussed above, there have been considerable advances~\cite{mirtaheri2021one,qin2020generative,jambor2021exploring,jiang2021metap} related to edge-level FSL. For example, ZSGAN~\cite{qin2020generative} studies zero-shot relation prediction by establishing the connection between text and knowledge graph with generative adversarial networks. Unlike node pair matching, P-INT~\cite{xu2021p} calculates the interactions of paths for relation prediction on knowledge graphs. Moreover, a recent study~\cite{jambor2021exploring} explores the limits of existing models for few-shot link prediction on knowledge graphs. 
\\
\textbf{Application.} Edge-level FSL methods have been applied to not only relation prediction problems on knowledge graphs, but also many other applications. For example, Meta-Graph~\cite{bose2019meta} investigates few-shot link prediction on different networks (e.g., biological network).
SEATLE~\cite{li2020few} and MetaHIN~\cite{lu2020meta} aim to tackle cold-start recommendation problems over graphs.

\subsection{Graph-level FSL}
Besides node-level and edge-level mining, graph-level learning is also significant to some application domains, such as bioinformatics and social network. Similar to the former two problems, the generation of labeled graph samples also involves scarcity and difficulty issues. For example, the collection of molecular graph labels for therapeutic activity often costs much money and resource~\cite{guo2021few}; some communities in social network (e.g., Reddit) only have a small number of sub-communities. In light of this, some studies have been proposed recently for solving graph-level problems with small labeled data. In this part, we summarize the latest development of graph-level FSLG. 
\\
\textbf{Metric-based Method.} Similar to MN-FSL, Metric-based Graph-level FSL (MG-FSL) focuses on computing distance between graph class prototype and query graph to predict labels of query graphs. That is, MG-FSL addresses the challenges of few-shot graph classification problem by combing graph-level GNN with ProNet. In particular, SuperClass~\cite{chauhan2019few} is the first work of MG-FSL for few-shot graph classification. Specifically, unlike the general ProNet which computes the average of support samples' embedding as class prototype, SuperClass employs a graph-level GNN (i.e., GIN~\cite{xu2018powerful}) to learn graph embedding, and then clusters graphs into different super-classes by computing prototype graphs from each class, followed by clustering the prototype graphs based on their spectral properties. Finally, the model is optimized using joint classification losses of both graph labels and super-class labels. More recently, different from the typical few-shot graph classification task performed on single domain data, MVG-Meta~\cite{hassani2022cross} develops a multi-view enhanced GIN to learn graph embedding for cross-domain few-shot graph classification, i.e., transferring meta-knowledge learned from one domain to another domain.  
\\
\textbf{Optimization-based Method.} Similar to ON-FSL and OE-FSL, Optimization-based Graph-level FSL (OG-FSL) leverages MAML to optimize few-shot graph classification model. In this way, OG-FSL is able to solve the challenges of few-shot graph classification problem. There are several studies of OG-FSL. Specifically, AS-MAML~\cite{ma2020adaptive} generates graph embedding by concatenating mean and max-pooling of all node embeddings encoded by GNN, and further leverages a reinforcement learning-based controller to adaptively control MAML for model optimization.
More recently, Meta-MGNN~\cite{guo2021few} and Pre-PAR~\cite{wang2021property} have been proposed to solve  molecular property prediction problem (i.e., molecular graph classification). Meta-MGNN takes each molecule as a graph and learns its embedding with graph-level GNN. It further introduces task weight to make MAML be aware of molecular property differences for better model optimization. Furthermore, Pre-PAR improves Meta-MGNN by modeling relational structure among different molecular properties, such that the limited labels can be effectively
propagated among similar molecules. 
\\
\textbf{Application.} As discussed above, besides general graph classification on different types of graphs (e.g., social network and biological network), existing graph-level FSL models have been applied to some other applications, such as molecular property prediction.  

\section{FSLG Future Research Directions}
FSLG is an emerging and fast-developing research topic. Although substantial progresses have been achieved, many challenges still remain. This opens up a number of avenues for future research directions. In this section, we identify and briefly discuss some of them. 
\begin{itemize}[leftmargin=*]
\item \textbf{Generalization and transferability.} Most of FSLG models excessively rely on labeled data and attempt to inherit a strong inductive bias for new tasks in the test phase. However, a distribution shift often exists between non-overlapping meta-training data and meta-testing data. Without supervision signals from ground-truth labels, the model may not learn an effective GNN for new classes of test data. This gap limits generalization and transferability of the meta-trained GNN. Fortunately, contrastive learning~\cite{chen2020simple,you2020graphcl} has emerged to alleviate the dependence on labeled data and learn label-irrelevant but transferable representation from unsupervised pretext tasks. Therefore, we may leverage contrastive learning to improve the generalization and transferability capability of current FSLG methods. 
\item \textbf{Explainability.} The previous FSLG studies target developing better models in performance while none of them has thought about model explanation. However, developing FSLG models with explainability is essential to improve model reliability and end-user trust.  For example, it is worth investigating and explaining which part of FSLG (e.g., GNN, meta-training, or meta-adaptation) is more significant to model performance, such that we can have a better guide for model design. In addition, we can develop information-based method~\cite{guan2019towards} to quantitatively explain FSLG model's capability. 

\item \textbf{Graph models for FSLG.} Though FSLG has been widely studied for certain types of graphs (e.g., plain graphs, attributed graphs), many other types of graphs such as signed graphs, multiplex graphs remain largely understudied in this filed. Meanwhile, the underlying GNN models adopted by existing FSLG work commonly follow the homophily principle, which cannot naturally adapt to heterophily graphs, where connected nodes are dissimilar. Hence, how to design principled graph models for graphs with different properties is also a promising research direction in the field of FSLG.


\item \textbf{Theoretical analysis of FSLG.} A recent work~\cite{jambor2021exploring} has empirically explored the limits of existing FSLG methods in relation prediction over knowledge graphs and challenged the implicit assumptions and inductive biases of prior work. To take a step further and unveil the profound foundation, theoretical analysis of FSLG, which has not been explored before, is necessary and important for us to better understand FSLG methods. In particular, a number of recent work~\cite{cao2019theoretical,du2020few,tripuraneni2021provable} related to FSL theory could serve as bases for this research direction. 
\item \textbf{Broader applications.} 
As discussed in this work, FSLG have been applied to not only general graph mining tasks at different granularity levels on various types of graphs but also some domain-specific applications (e.g., malware detection, illicit drug trafficker detection). Besides these studies, it is worth exploring the potential of FSLG to other application domains~\cite{mandal2022metalearning}, such as healthcare and social good. For example, we may develop FSLG model to capture patients' drug refill behavior (in which labels are limited) over prescription dispensing and refill data (modeled as graph), thus further performing early intervention of patients' abnormal behavior (e.g., opioid overdose), which is essential to their health. 
 
\end{itemize}

\section{Conclusion}
As two popular research topics in the AI community, graph representation learning and few-shot learning have laid the groundwork for a new promising research direction: \textbf{\underline{f}}ew-\textbf{\underline{s}}hot \textbf{\underline{l}}earning on \textbf{\underline{g}}raphs (FSLG), which has significance to various application domains. In this work, we first introduce major challenges and bases of FSLG. Then we comprehensively review existing studies of FSLG by systematically categorizing them into three parts for node-level, edge-level, and graph-level problems, respectively. Finally, we discuss several critical issues that should be solved and share our thoughts of future directions. We hope this review will serve as a useful reference for researchers and advance future work of FSLG.

\clearpage
\small
\bibliographystyle{named}
\bibliography{refs}

\end{document}